# Application of PSO, Artificial Bee Colony and Bacterial Foraging Optimization algorithms to economic load dispatch: An analysis

**Anant Baijal[1], Vikram Singh Chauhan[2] and T Jayabarathi[3]**

School of Electrical Engineering (SELECT), VIT University
Vellore, TN (632014), India

**Abstract**

This paper illustrates successful implementation of three evolutionary algorithms, namely- Particle Swarm Optimization (PSO), Artificial Bee Colony (ABC) and Bacterial Foraging Optimization (BFO) algorithms to economic load dispatch problem (ELD). Power output of each generating unit and optimum fuel cost obtained using all three algorithms have been compared. The results obtained show that ABC and BFO algorithms converge to optimal fuel cost with reduced computational time when compared to PSO for the two example problems considered.

***Keywords:*** *ABC, PSO, BFO, Economic Load dispatch, Evolutionary Algorithms*

## 1. Introduction

One of the conventional methods of solving ELD is lambda-iteration method but owing to tedious calculations and its inability to solve multi-modal and discontinuous problems, novel techniques have replaced it. Some of the techniques include Genetic Algorithm (GA), Dynamic Programming (DP), Evolutionary Programming (EP) [1-6] and neural networks. Swarm Optimization techniques are derived from Darwin's Evolutionary Theory of 'Survival of fittest'. In our paper we have discussed ELD problem solving by some swarm intelligence techniques namely, PSO, ABC and BFO algorithm. In past few years, lot of research work has been carried out on the above mentioned algorithms resulting in improvement of the efficacy of the same. Modifications of PSO include Improved PSO (IPSO), Hybrid Multi Agent (HMAPSO) [7] etc. while ABC variants are Honey Bee Mating Optimization (HBMO) [8], Bee Algorithm (BA) [9], Improved Bees Algorithm (IBA) [10] etc. and micro($\mu$)-BFO, Multi-objective Bacterial Foraging [11] etc. are some variants of the classical BFO algorithm. However, in this paper we have considered only classical versions of all three techniques.

The paper is organized as follows. Section 2 illustrates the ELD formulation Section 3, 4 & 5 elucidate briefly the PSO, ABC & BFO algorithms respectively. The numerical examples considered in this work and result analysis has been presented in section 6 followed by conclusion in section 7.

## 2. ELD Problem Formulation

The objective of ELD problem is to minimize the fuel cost while satisfying the constraints. In this work, we have considered constraints on power generation limits of each generator. The problem can be expressed as a quadratic function:

$$min\, f = \sum_{i=1}^{N} a_i P_i^2 + b_i P_i + c_i \qquad (1)$$

where *N* represents the number of generating units, $P_i$ is the power output (in MW) of the $i^{th}$ generator and *a,b* & *c* are the fuel cost coefficients.
Subject to:
Generators limit constraints:

$$P_{i\,min} \le P_i \le P_{i\,max} \qquad (2)$$

Equality constraint:

$$P_{demand} = \sum_{i=1}^{N} P_i \qquad (3)$$

The transmission and generator losses have been neglected.

## 3. The PSO Algorithm

### 3.1 Overview

Particle Swarm Optimization is an evolutionary computation technique developed by Eberheart & Kennedy [2] in 1995 and is based on bird flocking and fish schooling. PSO is a meta-heuristic technique as it makes few or no assumptions about the problem being optimized and can search very large spaces of candidate solutions. Over the decade, PSO has been proved to be one of the most promising algorithms for many intricate problems in engineering and sciences. Its simplicity and faster convergence make it an attractive algorithm to employ. The population is called swarm and the individuals are termed as particles. The word 'swarm' is inspired from jagged movement of particles in the problem region. The particles are assumed to be mass-less and volume-less.

### 3.2 Implementation

The dimension of the problem is determined by the number of generating units [12]. Then the current position of $i^{th}$ particle can be represented by $P_i = [P_{i1}, P_{i2}, P_{i3}....P_{iN}]$ where $P_i$ belongs to problem-space $S$. The particle flies with the current velocity given by $v = [v_{i1}, v_{i2}....v_{iN}]$ which is generated randomly in the range $[-v_{j\,max}\ v_{j\,max}]$. The objective function values are calculated using Eq. (1) and are set as $P_{best}$ values of the particles. The best value, based on individual fitness function is denoted by $G_{best}$ or the global best value of swarm. New velocity for every dimension in each particle is updated as in Eq. (4):

$$v_{ij}^{t+1} = w * v_{ij} + c_1 * rand * (P_{best} - P_{ij}^t) + c_2 * rand * (G_{best} - P_{ij}^t) \quad (4)$$

where $w$ is the weight vector whose value is to be suitably chosen, $c_1$ & $c_2$ are constants, $rand$ is a uniformly distributed random value between [0 1], $t$ represents iteration and $P_{ij}^t$ is the current position of the $j^{th}$ dimension in the $i^{th}$ particle. The new particle position is given by

$$P_{new} = P_{old} + v_{new} \quad (5)$$

When the stopping criteria are satisfied and there is no further improvement in the objective function, the position of the particles represented by $G_{best}$ gives the optimal dispatch.

## 4. ABC Algorithm

### 4.1 Overview

A number of algorithms based on bee-swarm have been developed, one of them being the Artificial Bee Colony algorithm which was proposed by Karaboga [13-14]. ABC is modeled on two processes: sending of bees to nectar (food source) and desertion of a food source. While in other swarm intelligence algorithms, the swarm represents the solution, in ABC the food source gives the solution while bees act as variation agents responsible for generating new sources of food. Three types of bees- employed, onlooker and scout bees, aid in reaching the optimal solution. This algorithm is very simple when compared to existing swarm algorithms.

### 4.2 Formulation

The initial position matrix is randomly generated by the action of scout bees while also satisfying the power balance constraints. Then the employed bees are sent onto the food source to determine the nectar content. Each bee is associated with only one food source. The action of employed bees, mathematically, is given by:

$$v_{ij} = x_{ij} + \phi(j)(x_{ij} - x_{kj}) \quad (6)$$

where $x_{ij}$ is the current solution in which the bee is located, $x_{kj}$ is a randomly generated food source and $\phi_{ij}$ is a real random number between [-1 1]. After search completion by all employed bees, the onlooker bees, using the roulette wheel probabilistic fitness function, communicate the food source information to other bees on the dance floor using waggle dancing, the frequency of which is dependent on the quality of food source. Hence more number of bees visit a profitable food source.

Now, if the fitness of a food source does not improve after a set limit, it is abandoned and replaced by a randomly generated food source by the scout bees; the act resembling negative feed back mechanism and fluctuation property of ABC. The best food source is remembered. This role-play cycle of scout-employed-onlooker-scout bees is repeated until the solution is optimized. The reader is referred to [14] for better understanding of ABC algorithm.

# 5. BFO Algorithm

## 5.1 Overview

The bacterial foraging optimization (BFO) algorithm is inspired from bio-mimicry of the e-coli bacteria and is a robust algorithm for non-gradient optimization solution, proposed in 2002 by Kevin M Passino [15]. It consists of four steps: chemo-taxis, swarming, reproduction & elimination-dispersal.

## 5.2 Implementation

The parameters initialized for run are: number of chemo-tactic steps (Nc), number of reproduction steps (Nre), number of elimination and dispersal steps (Ned), dispersal probability (Ped), number of bacteria (N) & swim length (Ns). An Ecoli can move in different ways: a 'run' shows movement in a p articular direction whereas a 'tumble' denotes change in direction. A tumble is represented by:

$$\theta^i(j+1,k,l) = \theta^i(j,k,l) + v(i)\phi(j) \quad (7)$$

where $\theta^i(j,k,l)$ represents $i^{th}$ bacterium in $j^{th}$ chemo-tactic, $k^{th}$ reproductive, $l^{th}$ elimination-dispersal step, $v(i)$ gives the step length and $\phi(j)$ is a unit length random direction given by:

$$\phi(j) = \frac{\Delta(i)}{\sqrt{\Delta(i)^T * \Delta(i)}} \quad (8)$$

At the end of specified chemo-tactic steps, the bacterium is evaluated and sorted in descending order of fitness. In the act of reproduction, the first half of bacteria is retained and duplicated while the other half is eliminated. Finally, bacteria are dispersed as per elimination and dispersal probability which helps hasten the process of optimization.

# 6. Numerical problems and Result analysis

The PSO, ABC and BFO algorithms were applied to obtain optimal fuel cost for two problems with data in Tables 1 and 2, each with three generating units.

Table 1: Generator Limits and Fuel Coefficients for Problem 1

$P_{demand} = 975MW$

| G | $P_{min}$ (MW) | $P_{max}$ (MW) | a | b | c |
|---|---|---|---|---|---|
| 1 | 200 | 450 | .004 | 5.3 | 500 |
| 2 | 150 | 350 | .006 | 5.5 | 400 |
| 3 | 100 | 325 | .009 | 5.8 | 200 |

Table 2: Generator Limits and Fuel Coefficients for Problem 2

$P_{demand} = 450MW$

| G | $P_{min}$ (MW) | $P_{max}$ (MW) | a | b | c |
|---|---|---|---|---|---|
| 1 | 100 | 600 | .0025 | 7.92 | 561 |
| 2 | 100 | 400 | .0019 | 7.85 | 310 |
| 3 | 50 | 200 | .0048 | 7.97 | 78 |

The optimal cost of first problem and second problem was found out to be 8237 $/hr and 4652 $/hr using lambda iteration method. While PSO took 50 iterations, ABC & BFOA took 5 and 10 iterations respectively to arrive at the optimal cost for the first problem. A total of 35, 4 & 11 iterations respectively were required to converge to the optimal solution for the second problem. It is to be noted that the number of iterations for BFO is actually the number of chemo-tactic steps and cannot be directly compared with the number of iterations for ABC & PSO.

Table 3: Comparison of Iterations and Computational Time (Problem 1)

| Algo. | P1 (MW) | P2 (MW) | P3 (MW) | Optimal Cost ($/hr) | Iterations | Computational Time (ms)* |
|---|---|---|---|---|---|---|
| PSO | 450 | 325 | 200 | 8237 | 50 | 7.80 |
| ABC | 450 | 325 | 200 | 8237 | 5 | 3.94 |
| BFO | 450 | 325 | 200 | 8237 | 10 | 3.89 |

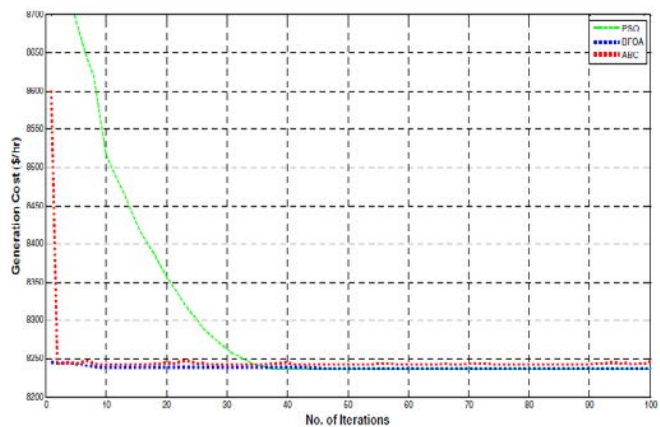

Fig. 1 Cost Function vs. number of iterations for Problem1

Table 4: Comparison of Iterations and Computational Time (Problem 1)

| Algo. | P1 (MW) | P2 (MW) | P3 (MW) | Optimal Cost ($/hr) | Iterations | Computational Time (ms)* |
|---|---|---|---|---|---|---|
| PSO | 206 | 184 | 60 | 4653 | 35 | 7.69 |
| ABC | 206 | 184 | 60 | 4653 | 4 | 3.91 |
| BFO | 206 | 183 | 61 | 4652 | 11 | 3.951 |

* The final computational time has been obtained by averaging the times for 100 runs.

The computational time using ABC and BFO is approximately 4ms and using PSO it is 8ms. The time is system dependent, in our case Intel Core 2 duo, T5800 @ 2 GHz. The convergence of the solutions to optimal fuel costs for the two examples considered here are shown in Fig.1 & Fig.2.

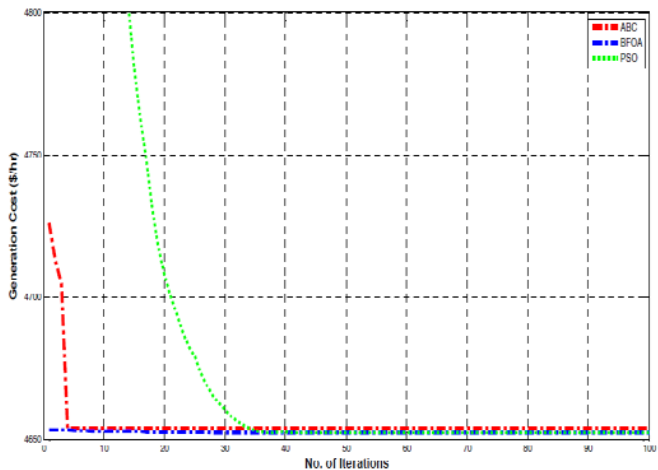

Fig. 2 Cost Function vs. number of iterations for Problem1

## 7. Conclusion

The application of PSO, ABC and BFO algorithms to constrained three-generator ELD problems has been successfully demonstrated with two examples. The test results show that ABC and BFO algorithms converge to optimal fuel cost with reduced computational time when compared to PSO. This is primarily due to the role of scout bees in ABC and the dispersal feature in BFO that introduce randomness during the optimization process resulting in significant computational time reduction. Further research work may focus on analyzing convergence rates by applying variants of the aforementioned algorithms and developing novel algorithms with higher efficacy.

**Acknowledgment**

The authors acknowledge VIT University, Vellore, for providing an opportunity to carry this work. The first and second authors also thank their colleagues Mr. Harsh Mathur & Mr. Jitendra Meel for assistance.